\documentclass{article}

\usepackage[final]{neurips_2021}

\usepackage[utf8]{inputenc} 
\usepackage[T1]{fontenc}    
\usepackage{hyperref}       
\usepackage{url}            
\usepackage{booktabs}       
\usepackage{amsfonts}       
\usepackage{nicefrac}       
\usepackage{microtype}      
\usepackage{xcolor}         
\usepackage{multirow}
\usepackage{graphicx}
\usepackage{amsmath}
\usepackage{caption}
\usepackage{subcaption}
\usepackage{tablefootnote}

\title{Prune Once for All: Sparse Pre-Trained Language Models}

\author{
  Ofir Zafrir \\
  Intel Labs, Israel \\
  \texttt{ofir.zafrir@intel.com}
  \And
  Ariel Larey \\
  Intel Labs, Israel \\
  \texttt{ariel.larey@intel.com}
  \And
  Guy Boudoukh \\
  Intel Labs, Israel \\
  \texttt{guy.boudoukh@intel.com}
  \And
  Haihao Shen \\
  Intel Corporation \\
  \texttt{haihao.shen@intel.com}
  \And
  Moshe Wasserblat \\
  Intel Labs, Israel \\
  \texttt{moshe.wasserblat@intel.com}
}

\begin{document}

\maketitle

\begin{abstract}
Transformer-based language models are applied to a wide range of applications in natural language processing.
However, they are inefficient and difficult to deploy.
In recent years, many compression algorithms have been proposed to increase the implementation efficiency of large Transformer-based models on target hardware.
In this work we present a new method for training sparse pre-trained Transformer language models by integrating weight pruning and model distillation.
These sparse pre-trained models can be used to transfer learning for a wide range of tasks while maintaining their sparsity pattern. 
We demonstrate our method with three known architectures to create sparse pre-trained BERT-Base, BERT-Large and DistilBERT.
We show how the compressed sparse pre-trained models we trained transfer their knowledge to five different downstream natural language tasks with minimal accuracy loss.
Moreover, we show how to further compress the sparse models' weights to 8bit precision using quantization-aware training.
For example, with our sparse pre-trained BERT-Large fine-tuned on SQuADv1.1 and quantized to 8bit we achieve a compression ratio of $40$X for the encoder with less than $1\%$ accuracy loss.
To the best of our knowledge, our results show the best compression-to-accuracy ratio for BERT-Base, BERT-Large, and DistilBERT.
\end{abstract}

\section{Introduction}
\label{sec:intro}
Transformer-based pre-trained language models (LM) such as BERT \citep{devlin2018bert} and RoBERTa \citep{liu2019roberta} have become the standard approach for a wide range of natural language processing (NLP) tasks.
Recently, we witness the emergence of models, larger by several orders of magnitude, such as GPT-2 \citep{radford2019gpt2}, T-NLG \citep{rosset2020tnlg}, GPT-3 \citep{brown2020gpt3}, and Switch-C \citep{fedus2021switch}. 
These models advance the state-of-the-art results in several NLP tasks such as question answering and text classification.
However, this trend toward bigger models raises several concerns.
As the computational and memory resources required to run inference increase with the model's size, it becomes very expensive and challenging to deploy these models in production environments and on edge devices.
Moreover, these large amounts of computational resources incur a steep environmental cost \citep{strubell2019green}.

Model compression of large LM is a growing field of study as a result of these concerns.
Weight pruning is a compression method that has been shown to be very effective at reducing the memory footprint of a model \citep{han2015imp, zhu2017prune}.
However, weight pruning of large Transformer-based LMs to high sparsity ratios requires specialized pruning methods \citep{sanh2020movement, chen2020lottery, gordon2020pre-train, lagunas2021block}.
Moreover, most of the pruning methods require task specific modifications and tuning to produce quality results.

\citet{gordon2020pre-train} found that, in terms of accuracy, it does not matter whether BERT is pruned during the pre-training phase or during the transfer learning phase.
This suggests that a LM can be pruned once during pre-training and then fine-tuned to any downstream task without task-specific tuning.

In this paper, we present a new method, Prune Once for All (Prune OFA), that leverages weight pruning and model distillation to produce pre-trained Transformer-based language models with a high sparsity ratio.
We apply our method to BERT-Base, BERT-Large and DistilBERT \citep{sanh2019distilbert} to produce sparse pre-trained models for these model architectures.
We then show how these sparse models can be fine-tuned to produce task-specific sparse models with minimal accuracy loss for SQuADv1.1 \citep{rajpurkar2016squad} as well as for four tasks from the GLUE Benchmark \citep{wang2018glue}.
We also show that it is possible to further compress the models using quantization-aware training to achieve state-of-the-art results in terms of compression-to-accuracy ratio.

The main contributions of this work are threefold: 
1) We introduce a new architecture-agnostic method of training sparse pre-trained language models.
2) We demonstrate how to fine-tune these sparse models on downstream tasks to create sparse and quantized models, removing the burden of pruning and tuning for a specific language task.
3) We publish our compression research library with example scripts to reproduce our work for other architectures, along with our sparse pre-trained models presented in this paper.

\section{Related work}
\label{sec:related}
Large language models are over-parameterized and difficult to deploy. Therefore, the problem of compressing these models with minimum accuracy loss for downstream tasks is widely explored. 
\citet{sanh2020movement} suggests the Movement Pruning method designed especially for transfer learning.
Neural Magic implements Gradual Magnitude Pruning.\footnote{\url{https://github.com/neuralmagic/sparseml/tree/main/integrations/huggingface-transformers}}
Both methods suggest pruning BERT-Base while fine-tuning to downstream tasks paired with model distillation, and present results showing $90\%$ sparsity for several tasks.
However, both methods require a long fine-tuning time as well as tuning pruning related hyper-parameters for every task. 
Our method, on the other hand, requires no tuning of special pruning hyper-parameters  per task because we prune the model once for all tasks.
Furthermore, we present better or comparable results at a much lower computation budget at the transfer learning phase.
\citet{gordon2020pre-train} explored the effect of weight pruning during transfer learning and concluded that pruning BERT-Base at the pre-training phase does not degrade the performance of the model compared to pruning at fine-tuning.
We improve upon the suggested method and present better results at a much higher sparsity ratio.
\citet{chen2020lottery} explored the Lottery Ticket Hypothesis \citep{frankle2018lth} for BERT pre-trained models.
More specifically, they analyzed the possibility of finding winning tickets in a BERT-Base pre-trained model that transfer to other downstream tasks.
The authors concluded that winning tickets found while pre-training on a Masked-LM task, transfer well to other downstream tasks.
\citet{lagunas2021block} presented a structured pruning method, removing rows, columns and attention heads, while achieving less than $1\%$ loss in F1 for a BERT architecture on SQuADv1.1.
\citet{mishra2021nvidia-pre-train} performed structured 2:4 pruning on BERT while further pre-training BERT;
The method produced a $50\%$ sparse model which can be fine-tuned without accuracy loss.
\citet{michel2019heads} explored the significance of each head in the multi-head attention mechanism of BERT and presented a method for pruning attention heads with their associated weights.

Other works propose knowledge distillation to compress Transformer models to a smaller dense counter part that can be tuned to downstream tasks \citep{sanh2019distilbert, jiao2019tinybert, sun2020mobilebert}.
Quantization of Transformer-based language models is also a well known method for compression.
\citet{shen2020qbert} proposes a method to quantize BERT at a different bit-width per layer.
Other works implement quantization-aware training to quantize BERT to 8bits \citep{kim2021ibert, zafrir2019q8bert}.
\citet{zhang2020ternarybert} created a method of producing a ternary weight BERT.
\citet{kim2020fastformers} presented a compression pipeline for Transformer models that includes model distillation, quantization and head pruning.

\section{Weight pruning}
\label{sec:weight-pruning}
Weight pruning is the process of forcing some of the neural network's weights to zero.
Weight pruning can be either unstructured where individual weights are pruned, or structured where structured groups of weights are pruned, e.g. blocks, channels, layers.
Weight pruning results in sparse neural networks that reduce the computation and the memory footprint of the trained model.

In this paper we focus on unstructured weight pruning.
\citet{zhu2017prune} presented a method of Gradual Magnitude Pruning (GMP) to gradually prune weights with low magnitude during training.
During training, every $f$ steps the lowest magnitude weights are pruned until reaching the temporal sparsity ratio $s_t$ for time step $t$, defined by
\begin{equation}
\label{eq:pruning-sched}
    s_t = s_f + \left(s_i - s_f\right)\left(1 - \frac{t - t_s}{t_e - t_s} \right)^3
\end{equation}
where $s_i$ and $s_f$ are the initial and final sparsity ratios, and $t_s$ and $t_e$ are the pruning start and end time steps.

In a recent paper, \citet{renda2020rewind} presented a pruning algorithm based on IMP (Iterative Magnitude Pruning) \citep{han2015imp} and Learning Rate Rewinding (LRR).
IMP consist of two steps: prune a portion of the model and continue fine-tuning it to recover from the induced pruning error.
These two steps are repeated until the desired sparsity ratio is achieved.
In LRR, the learning rate scheduler is rewound to its state before the pruning step at the beginning of the fine-tune step.
We propose to incorporate the principle of learning rate rewinding into GMP by rewinding the learning rate scheduler to its state at time $t_s$ every $f$ steps.
After $t_e$ the scheduler continues with its original setting until training ends.
Appendix~\ref{app:lrr-visual} visualizes how LRR combined with GMP modifies the learning rate scheduler.

\section{Knowledge distillation}
\label{sec:kd}
Knowledge distillation, introduced by \citet{hinton2015distilling}, is the process of training a student network to reproduce the behavior of a teacher model. When distillation is used to fit the predictions of the teacher model, soft cross-entropy loss between the student and the teacher soft probabilities is computed as follows:
\begin{equation}
\label{eq:distillation loss}
    \mathcal{L}_{kd} = -\sum_{i}{t_i \cdot \log\left(s_i\right)}
\end{equation}
where $s_i$ is the soft-probability estimated by the student, and $t_i$ is its corresponding soft-probability estimated by the teacher for the same input sample. The soft probabilities are calculated using a \texttt{softmax} function with temperature $T$.

Commonly, the teacher is a large model that achieves high performance, and the student is based on a smaller architecture. 
In this paper, we propose to leverage the model distillation method for the pruning process. 
We focus on an approach where both teacher and student share the same architecture, but differ in their sparsity ratio. 
In this case, the teacher is a dense model that was trained on a target task, and the student is a model with a fixed sparsity or one undergoing pruning. 
Distillation-during-pruning can be applied to language models during both the pre-training and fine-tuning phases. 
In the pre-training phase, the teacher is a pre-trained language model, and in the fine-tuning phase, the teacher is a language model fine-tuned to a target task. 

\section{Prune Once for All}
\label{sec:sparse-pre-train-models}
\begin{figure}[t]
\centering
	\includegraphics[width=\textwidth]{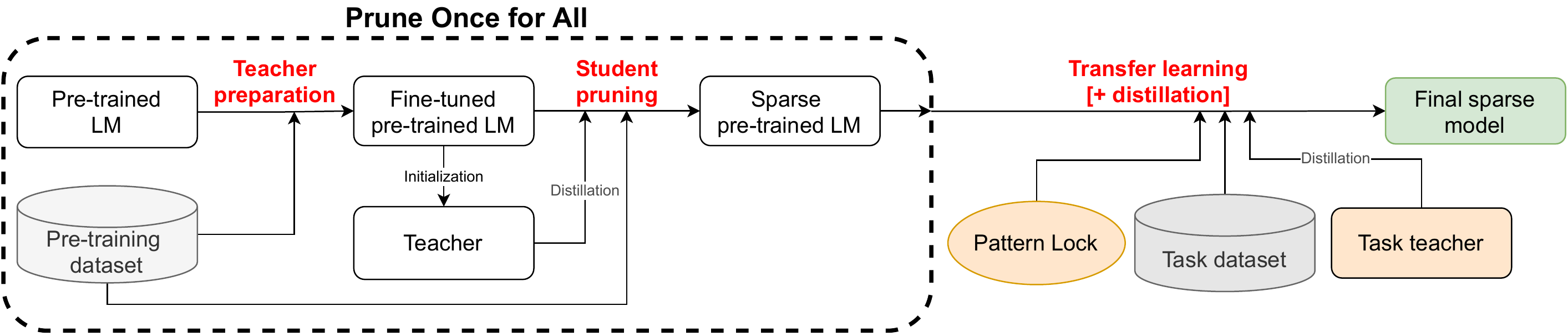}
    \caption{Prune OFA method}
    \label{fig:prune-once-for-all}
\end{figure}
The notion of pruning language models such as BERT \citep{devlin2018bert} while pre-training has already been explored by \citet{chen2020lottery} and \citet{gordon2020pre-train}.
However, fine-tuning the sparse model for a specific language task resulted in either poor results or a low sparsity ratio.
In this section we will introduce our novel method, Prune OFA, for creating sparse pre-trained language models that can be later fine-tuned to downstream tasks with minimal accuracy loss at high sparsity ratios.
A visualization of our method is presented in Figure~\ref{fig:prune-once-for-all}.
The method takes as its input a pre-trained language model and outputs a sparse language model of the same architecture.
The method consists of two steps, teacher preparation and student pruning. The sparse pre-trained model we trained is the model we use for transfer learning while maintaining its sparsity pattern.
We call the method Prune Once for All since we show how to fine-tune the sparse pre-trained models for several language tasks while we prune the pre-trained model only once.

\paragraph{Teacher preparation}
The first step of Prune OFA is to obtain a model optimized on the pre-training dataset for some pre-training task with objective $\mathcal{L_{PT}}$ as shown in Figure~\ref{fig:prune-once-for-all}.\footnote{For example, the pre-training task for BERT-Base is masked language-modeling combined with next sentence prediction.}
The same dataset will be used for pruning the student in the next step.
This model will initialize the student and teacher models in the student pruning step.

\paragraph{Student pruning}
A student model is initialized from the teacher prepared in the teacher preparation step.
The student is then fine-tuned on a linear combination of the pre-training task, from the teacher preparation step, and the knowledge  distillation objective $\mathcal{L}_{kd}$:
\begin{equation}
    \mathcal{L} = \lambda_{PT} \mathcal{L}_{PT} + \lambda_{kd} \mathcal{L}_{kd}
\end{equation}
while being pruned with GMP + LRR methods.
The output model of this process is a sparse pre-trained LM that can be used without additional pruning for transfer learning to produce sparse models for a specific downstream task.

\paragraph{Pattern-lock}
\label{sec:method-transfer}
We wish to keep the sparsity pattern of the sparse pre-trained model created by Prune OFA in place during the fine-tuning process.
We propose a method called pattern-lock that prevents the zeros found in the model from changing while training the model.
Pattern-lock is described in more details in Appendix~\ref{app:pattern-lock}.

\section{Experimental setup}
\label{sec:exp-setup}

\paragraph{Datasets}
\label{sec:datasets}
We use the English Wikipedia dataset (2500M words) for training the models on the pre-training task.
We split the data into train ($95\%$) and validation ($5\%$) sets.
Both sets are pre-processed as described in the models' original papers \citep{devlin2018bert, sanh2019distilbert}.
We process the data to use the maximum sequence length allowed by the models, however, we allow shorter sequences at a probability of $0.1$.
We evaluate our sparse pre-trained models on several common benchmarks for transfer learning; a question answering task, SQuADv1.1 containing 89K training examples \citep{rajpurkar2016squad}, and the following text classification tasks from the GLUE Benchmark: MNLI, QQP, QNLI and SST-2 containing 393K, 364K, 105K, and 67K training examples respectively \citep{wang2018glue, williams2018mnli, iyer2017qqp, socher2013sst2}.

\paragraph{Applying Prune Once for All}
\label{sec:apply-pofa}
We showcase our method by applying Prune OFA on three different architectures of different sizes; BERT-Base, BERT-Large and DistilBERT.
Since we don't have the original processed training data used to train BERT-Base, BERT-Large and DistilBERT we run an additional step to fine-tune the pre-trained models using the processed training data we prepared.
Next, we execute the student pruning step to obtain our sparse pre-trained models.
We prune BERT-Base and DistilBERT to $\left\{85\%, 90\%\right\}$ sparsity ratios and BERT-Large to a $90\%$ sparsity ratio.
Pruning is applied to all Linear layers in the Transformer encoder including the pooler layer if it exists.
Exact hyper-parameters and additional details are summarized in Appendix~\ref{app:reproduce}

\paragraph{Transfer learning}
After creating our sparse pre-trained models we fine-tune them to the following NLP tasks: SQuADv1.1, QNLI, MNLI, SST-2 and QQP.
We use default hyper-parameters for each task and conduct a grid search for learning rate, weight decay, warmup ratio and number of training epochs hyper-parameters.
For each task we report the mean of two different runs with different seeds that achieved the best result on the task's development set.
We further improve the results of our sparse models by integrating knowledge distillation.
For each task and model, we create a task teacher based on the original dense pre-trained model fine-tuned to the task.
For SQuADv1.1 and QQP we report the result that maximizes F1, and for MNLI we report the result that maximizes the mismatched accuracy.
For exact hyper-parameters and additional details see Appendix~\ref{app:reproduce}.

\paragraph{Comparison with fine-tune pruning}
We compare our Prune OFA method with fine-tune pruning where we prune the dense pre-trained model during fine-tuning to a downstream task. 
For that purpose, we implement GMP pruning coupled with knowledge distillation and run experiments using the same teacher and hyper-parameters used in the Prune OFA transfer learning experiments.

\paragraph{Quantization}
\label{sec:quantization}
We implemented quantization-aware training similar to Q8BERT \citep{zafrir2019q8bert}.
For details on the differences between our method and Q8BERT see Appendix~\ref{app:quantization}.
For each task, we pick the best-performing model for this task and perform quantization-aware training on it.
We use slightly different hyper-parameters for this training session as described in Appendix~\ref{app:hyper-params}.
We report the mean of two different runs with different seeds that achieved the best result.

\section{Results}
\label{sec:results}
\begin{table}[t]
\centering
\caption{Prune OFA BERT-Base results compared to other pruning methods}
\label{tab:bert-base-compare}
\resizebox{\textwidth}{!}{%
\begin{tabular}{@{}lcccccccccc@{}}
\toprule
\multirow{2}{*}{Model} & \multicolumn{1}{l}{\multirow{2}{*}{Sparsity}} & \multicolumn{1}{l}{\multirow{2}{*}{\begin{tabular}[c]{@{}l@{}}Transfer\\ with KD\end{tabular}}} & \multicolumn{2}{c}{SQuAD} & \multicolumn{2}{c}{MNLI (m/mm)} & SST-2 & QNLI & \multicolumn{2}{c}{QQP} \\
 & \multicolumn{1}{l}{} & \multicolumn{1}{l}{} & EM & F1 & Acc & Acc & Acc & Acc & Acc & F1 \\ \midrule
Reference & 0\% &  & 80.80 & 88.50 & 84.06 & 84.51 & 92.13 & 91.16 & 91.20 & 88.13 \\ \midrule
\citet{chen2020lottery} & 70\% &  & N/A & 86.54 & \textbf{82.59} & N/A & \textbf{91.86} & 89.44 & 90.03 & N/A \\
\citet{gordon2020pre-train} & 80\% &  & N/A & N/A & 75.90 & N/A & 88.10 & 85.30 & 86.90 & N/A \\
Prune OFA & 85\% &  & \textbf{78.59} & \textbf{86.63} & 81.67 & \textbf{82.53} & 91.34 & \textbf{89.95} & \textbf{90.69} & \textbf{87.41} \\ \midrule
Fine-tune pruning & \multirow{2}{*}{85\%} & + & 78.00 & 86.16 & 82.45 & 83.05 & 88.82 & 87.79 & 90.87 & 87.65 \\
Prune OFA &  & + & \textbf{81.10} & \textbf{88.42} & \textbf{82.71} & \textbf{83.67} & \textbf{91.46} & \textbf{90.34} & \textbf{91.15} & \textbf{88.00} \\ \midrule
Prune OFA +QAT & 85\% & + & 80.84 & 88.24 & 81.40 & 82.51 & 91.46 & 89.76 & 91.09 & 88.01 \\ \midrule
Neural Magic\tablefootnote{Results taken from Neural Magic's sparse model zoo: \url{https://sparsezoo.neuralmagic.com/}} & \multirow{3}{*}{90\%} & + & 79.40 & 87.20 & N/A & N/A & N/A & N/A & N/A & N/A \\
\citet{sanh2020movement} &  & + & 76.60 & 84.90 & 81.20 & 81.80 & N/A & N/A & 90.20 & 86.80 \\
Prune OFA &  & + & \textbf{79.83} & \textbf{87.25} & \textbf{81.45} & \textbf{82.43} & 90.88 & 89.07 & \textbf{90.93} & \textbf{87.72} \\ \bottomrule
\end{tabular}%
}
\end{table}

\begin{table}[t]
\centering
\caption{Prune OFA BERT-Large results}
\label{tab:bert-large-results}
\resizebox{\textwidth}{!}{%
\begin{tabular}{@{}lccccccccc@{}}
\toprule
\multirow{2}{*}{Model} & \multicolumn{1}{l}{\multirow{2}{*}{Sparsity}} & \multicolumn{2}{c}{SQuAD} & \multicolumn{2}{c}{MNLI (m/mm)} & SST-2 & QNLI & \multicolumn{2}{c}{QQP} \\
 & \multicolumn{1}{l}{} & EM & F1 & Acc & Acc & Acc & Acc & Acc & F1 \\ \midrule
Reference & 0\% & 83.99 & 90.93 & 86.39 & 86.58 & 93.54 & 92.42 & 91.59 & 88.67 \\
Prune OFA & 90\% & 83.35 & 90.20 & 83.74 & 84.20 & 92.95 & 91.39 & 91.48 & 88.43 \\
Prune OFA + QAT & 90\% & 83.22 & 90.02 & 83.47 & 84.08 & 92.72 & 91.45 & 91.41 & 88.36 \\ \bottomrule
\end{tabular}%
}
\end{table}

\begin{table}[t]
\centering
\caption{Prune OFA DistilBERT results compared to fine-tune pruning}
\label{tab:distilbert-results}
\resizebox{\textwidth}{!}{%
\begin{tabular}{@{}lccccccccc@{}}
\toprule
\multirow{2}{*}{Model} & \multicolumn{1}{l}{\multirow{2}{*}{Sparsity}} & \multicolumn{2}{c}{SQuAD} & \multicolumn{2}{c}{MNLI (m/mm)} & SST-2 & QNLI & \multicolumn{2}{c}{QQP} \\
 & \multicolumn{1}{l}{} & EM & F1 & Acc & Acc & Acc & Acc & Acc & F1 \\ \midrule
Reference & 0\% & 77.70 & 85.80 & 82.20 & N/A & 91.30 & 89.20 & N/A & 88.50 \\ \midrule
Fine-tune pruning & \multirow{2}{*}{85\%} & 76.16 & 84.55 & 81.22 & 81.92 & 88.88 & 86.60 & 90.18 & 86.80 \\
Prune OFA &  & \textbf{78.10} & \textbf{85.82} & \textbf{81.35} & \textbf{82.03} & \textbf{90.60} & \textbf{88.31} & \textbf{90.29} & \textbf{86.97} \\ \midrule
Prune OFA +QAT & 85\% & 77.03 & 85.13 & 80.66 & 81.14 & 88.93 & 87.97 & 90.22 & 86.92 \\ \midrule
Fine-tune pruning & \multirow{2}{*}{90\%} & 74.63 & 83.42 & 80.47 & 81.32 & 88.25 & 84.91 & 89.97 & 86.57 \\
Prune OFA &  & \textbf{76.91} & \textbf{84.82} & \textbf{80.68} & \textbf{81.47} & \textbf{90.02} & \textbf{87.66} & \textbf{90.05} & \textbf{86.67} \\ \midrule
Prune OFA +QAT & 90\% & 75.62 & 83.87 & 78.80 & 80.40 & 88.47 & 87.20 & 89.97 & 86.63 \\ \bottomrule
\end{tabular}%
}
\end{table}

In Table~\ref{tab:bert-base-compare} we present our experimental results for pruning BERT-Base to a $85\%$ and $90\%$ sparsity ratio using Prune OFA.
We also present results of other pruning methods applied to BERT-Base as well as results of the fine-tune pruning experiments we conducted.
Results not marked in the column Transfer with KD do not use model distillation in the transfer learning phase.
The best result in each category is marked with bold font.
We observe that our method achieves better results than other previous pruning works while pre-training at a higher sparsity ratio.
When comparing our Prune OFA method against other fine-tune pruning methods, we observe that our method produces the best results at $85\%$ and $90\%$ sparsity ratios.
Moreover, we show accuracy degradation lower than $1\%$ relative to the results of the dense pre-trained model at $85\%$ sparsity with the exception of the MNLI-matched benchmark.
Note that for MNLI, the reported results were selected based on the best model's mismatched accuracy found in our grid-search; when searching for the best matched result we reduce the accuracy gap to $\sim 1\%$ accuracy loss at the expense of increased accuracy loss for mismatched: $83.09$/$83.36$ (m/mm).

The results for pruning BERT-Large to a $90\%$ sparsity ratio are presented in Table~\ref{tab:bert-large-results}.
These results fall within the range of $1\%$ accuracy loss for all tasks but the MNLI task.
We conclude that the $90\%$ sparse BERT-Large ($30.2$M non-zero parameters) model we trained has better accuracy in comparison to dense BERT-Base ($85$M non-zero parameters).

Our results for pruning DistilBERT to a $85\%$ and $90\%$ sparsity ratio are presented in Table~\ref{tab:distilbert-results} with our results for the fine-tune pruning experiments we conducted.
In both sparsity ratios our method produces better accuracy results compared to fine-tune pruning (the best result in each category is marked with bold font).
Furthermore, at the 85\% sparsity ratio our results are within the range of $1\%$ relative accuracy loss in all tasks but QQP.

Tables~\ref{tab:bert-base-compare}, \ref{tab:bert-large-results} and \ref{tab:distilbert-results} present quantization results, designated with a +QAT suffix.
Applying quantization-aware training on our resultant sparse models decreases the accuracy of the model further by an average of $0.67\%$ relative to the full precision model's accuracy.
The results for the $85\%$ sparse model +QAT are better than for the $90\%$ sparse model with full precision in all the tasks for BERT-Base and in 3/5 tasks for DistilBERT.
Furthermore, the $85\%$ sparse and quantized model are smaller than the $90\%$ sparse model by a factor of $0.375$.

An ablation study was conducted to test how each component of the Prune OFA method affects the ability of the pre-trained model to transfer its knowledge to downstream tasks, as described in Appendix~\ref{app:ablation}.

\section{Conclusion and future work}
\label{sec:conclusions}
We introduced Prune OFA, an architecture-agnostic method for producing sparse pre-trained language models.
We also showed how these sparse models can be used to obtain fine-tuned sparse models without the burden of task-specific pruning.
Our results suggest that using these sparse pre-trained models for transfer learning produces results with minimal performance degradation loss w.r.t their dense counterpart for a variety of NLP tasks.
We further demonstrated that integrating quantization can lead to more efficient sparse and quantized models at a small cost to the model's accuracy.

A possible direction for future research is to explore whether a large and sparse pre-trained model is better at capturing and transferring natural language knowledge than a smaller dense model of the same architecture with similar non-zero parameters count.

We hope that the release of our code and sparse pre-trained models to the community will help develop more efficient models.

\section{Acknowledgements}
We are grateful to Ella Charlaix of HuggingFace for her fruitful comments and corrections.

\bibliographystyle{abbrvnat}
\bibliography{references}

\appendix
\section{Ablation study}
\label{app:ablation}
\begin{table}[t]
\centering
\caption{Prune OFA 85\% sparse BERT-Base ablation study results}
\label{tab:ablation}
\begin{tabular}{@{}cccccccc@{}}
\toprule
\multicolumn{1}{l}{\multirow{2}{*}{\begin{tabular}[c]{@{}l@{}}Teacher\\ preparation\end{tabular}}} & \multicolumn{1}{l}{\multirow{2}{*}{LRR}} & \multicolumn{1}{l}{\multirow{2}{*}{\begin{tabular}[c]{@{}l@{}}Pre-train\\ distillation\end{tabular}}} & \multicolumn{1}{l}{\multirow{2}{*}{\begin{tabular}[c]{@{}l@{}}Transfer\\ distillation\end{tabular}}} & \multicolumn{2}{c}{SQuAD} & \multicolumn{2}{c}{MNLI (m/mm)} \\
\multicolumn{1}{l}{} & \multicolumn{1}{l}{} & \multicolumn{1}{l}{} & \multicolumn{1}{l}{} & EM & F1 & Acc & Acc \\ \midrule
 &  &  &  & 78.11 & 86.13 & 81.14 & 81.74 \\
+ &  &  &  & 78.00 & 86.31 & 81.22 & 82.01 \\
+ & + &  &  & 78.41 & 86.51 & 81.39 & 82.01 \\
+ &  & + &  & 78.30 & 86.41 & 81.57 & 82.13 \\
+ & + & + &  & 78.59 & 86.63 & 81.67 & 82.53 \\ \midrule
+ &  &  & + & 80.77 & 88.08 & 82.20 & 82.83 \\
+ & + & + & + & 81.10 & 88.42 & 82.71 & 83.67 \\ \bottomrule
\end{tabular}%
\end{table}

In this section we analyze how each step of the Prune OFA method affects the final results.
We compare the models in the same fashion as in Section~\ref{sec:results}, by comparing the different  results of the sparse pre-trained models on downstream tasks.
In the ablation study we focus on BERT-Base pruned to 85\% fine-tuned to SQuADv1.1 and MNLI.
All the results from the ablation study are present in Table~\ref{tab:ablation}.

\paragraph{Teacher preparation}
The teacher preparation step is only done in the case the original processed training data of the pre-trained model is not available.
Since our objective is to prune the model it is always better to start from a model that is better optimized to the data used for pruning, hence the teacher preparation step.
To measure the effect of the teacher preparation step we prune two models, a model that uses BERT-Base pre-trained model as initialization, and a model that uses the output of the teacher preparation step as initialization.
Then, we fine-tune them both to SQuADv1.1 and MNLI tasks and compare their results.
We see notable improvement when executing with the teacher preparation step in both tasks.

\paragraph{Student pruning}
We compare the results of a model pruned with LRR to a model that was pruned without LRR, meaning the learning rate schedule remained the default linear decay with warmup schedule.
For SQuADv1.1 we observe a significant improvement in both benchmarks.
However, in MNLI case we don't see any improvement in the mismatched accuracy which we try to maximize, but there is a significant improvement in the matched accuracy.
We observe that applying knowledge distillation during the student pruning step improves both tasks results.
Knowledge distillation seems less significant in SQuADv1.1 case and more significant in MNLI case.
In addition, we see that combining LRR and knowledge distillation achieves better results than either method separately.
We conclude that applying LRR while pruning improves fine-tuning results and therefore a crucial part of our algorithm.

\paragraph{Transfer learning with knowledge distillation}
We saw that using knowledge distillation while fine-tuning to downstream tasks improves the results significantly.
We test whether our method still improves accuracy results of sparse models when fine-tuned with model distillation.
From the results at the bottom of Table~\ref{tab:ablation} we deduce that our method is orthogonal to knowledge distillation while fine-tuning and improves the accuracy results of both tasks further.

\section{Pattern-lock details}
\label{app:pattern-lock}
Following is a detailed description of the Pattern-lock method used when fine-tuning our sparse pre-trained models.
Before training, Pattern-lock method initializes a mask $M^l$ for each sparse layer $l$ with weight $W^l$, representing the layer's sparsity pattern.
\begin{equation}
\label{eq:mask-init}
    M^l_{uv} = 
    \begin{cases}
    1 & W^l_{uv} \ne 0 \\
    0 & W^l_{uv} = 0
    \end{cases}
\end{equation}
Then, while training, the loss $\mathcal{L}$ gradient w.r.t the weights is modified to
\begin{equation}
    \overline{\frac{\partial \mathcal{L}}{\partial W^l_{uv}}} = 
    \begin{cases}
    \frac{\partial \mathcal{L}}{\partial W^l_{uv}} & M^l_{uv} = 1 \\
    0 & M^l_{uv} = 0
    \end{cases}
\end{equation}
ensuring that a weight that was initially $0$ will stay $0$ through-out fine-tuning.

\section{Visualization of Learning Rate Rewinding with Gradual Magnitude Pruning}
\label{app:lrr-visual}
\begin{figure}[t]
    \centering
    \begin{subfigure}[b]{0.49\textwidth}
        \centering
        \includegraphics[width=\textwidth]{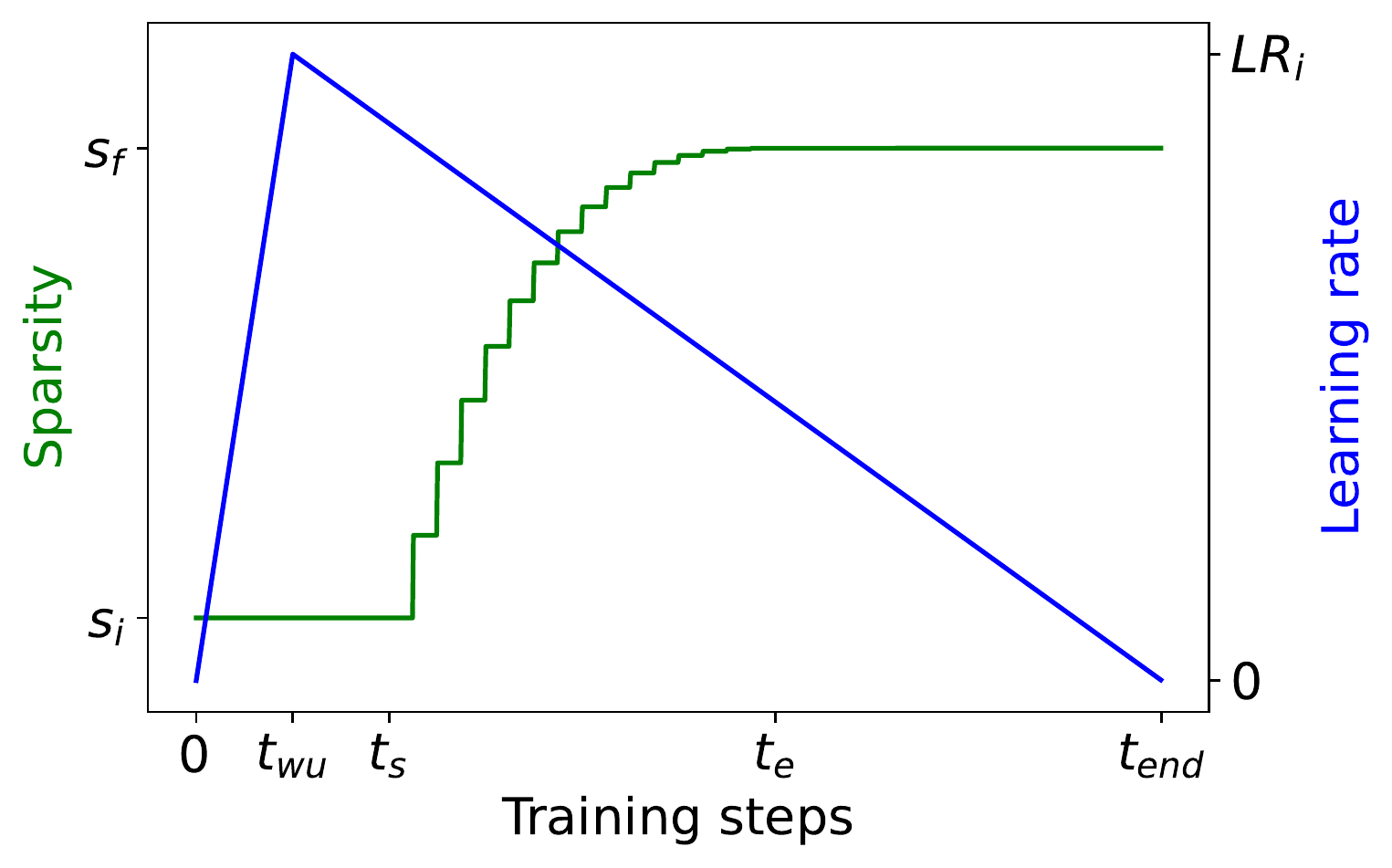}
        \caption{}
        \label{fig:lr_norewind_sparsity_sched}
    \end{subfigure}
    \hfill
    \begin{subfigure}[b]{0.49\textwidth}
        \centering
        \includegraphics[width=\textwidth]{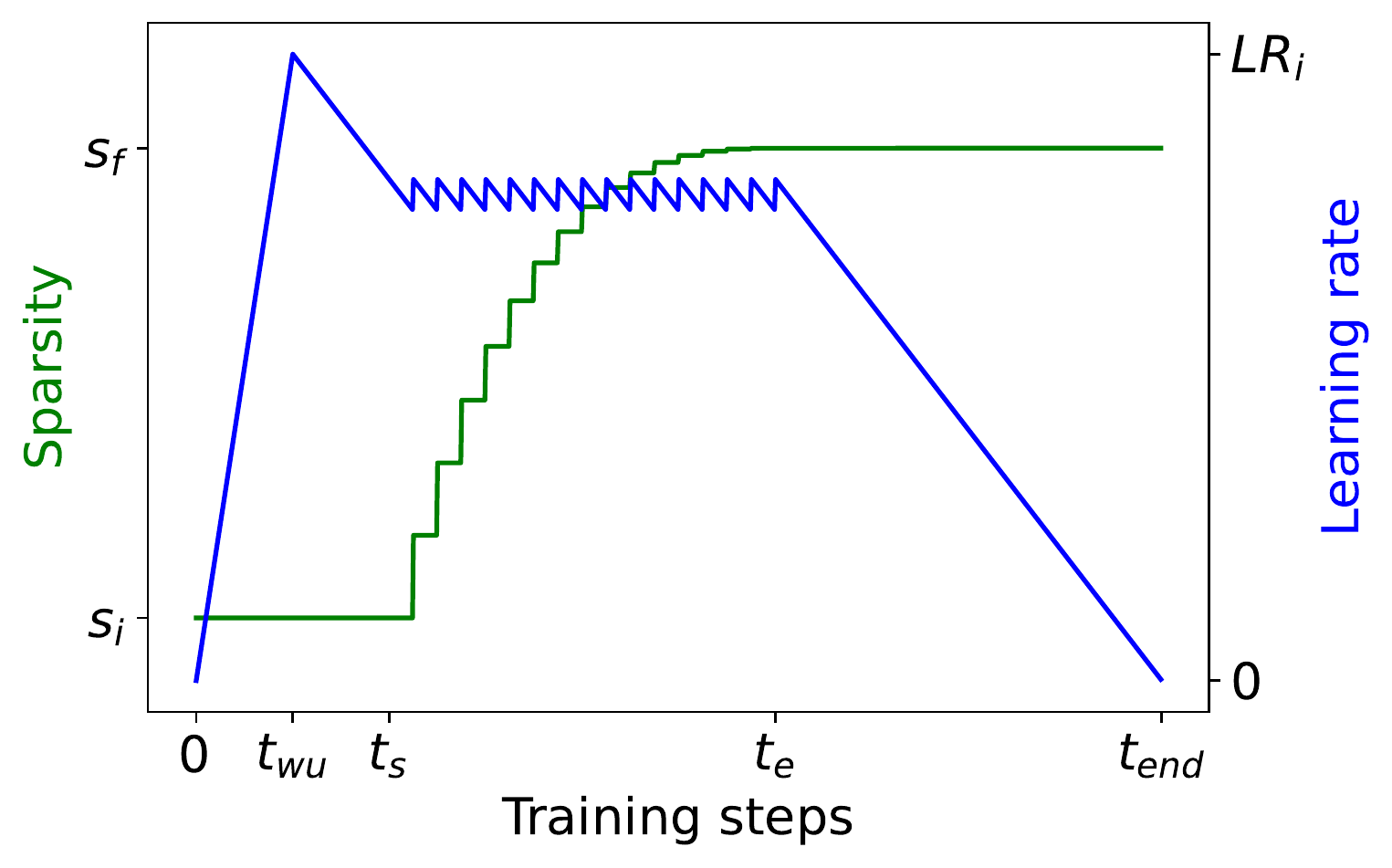}
        \caption{}
        \label{fig:lr_rewind_sparsity_sched}
    \end{subfigure}
    \caption{Learning rate and sparsity scheduler. Both figures show a linear decay learning rate scheduler with $t_{wu}$ warmup steps against a sparsity scheduler defined by Equation~\ref{eq:pruning-sched}. (a) learning scheduler without rewinding. (b) learning scheduler with rewinding}
    \label{fig:lr_sparsity_sched}
\end{figure}
Figure~\ref{fig:lr_sparsity_sched} demonstrates how a linear decay learning rate scheduler with warmup is modified with LRR against the same scheduler without LRR.

\section{Quantization method differences from Q8BERT}
\label{app:quantization}
We have implemented our own version of quantization-aware training which is similar to Q8BERT with the following differences:
1) Activations are quantized using asymmetric quantization instead of symmetric quantization.
2) Embedding vectors are not quantized and represented in full precision.
3) Models are quantized after fine-tuning to a downstream task in a seperate learning session.

\section{Reproducibility}
\label{app:reproduce}
\subsection{Implementation}
Our Prune OFA method, GMP, model distillation and quantization-aware training are implemented in our Model Compression Research Package using \texttt{PyTorch} \citep{paszke2019pytorch}.\footnote{\url{https://github.com/IntelLabs/Model-Compression-Research-Package}}
Our library offers several architecture agnostic pruning and other compression methods that can be plugged into any training session with a few lines of code.
We invite the researches community to use our library to accelerate their research in pruning and neural networks compression.

We use the \texttt{HuggingFace/transformers} library and the available example scripts to train our Transformer-based models \citep{wolf-etal-2020-transformers}.
We have modified the example scripts to include our methods and make them available in our library's examples.

All the datasets mentioned in the paper are downloaded and processed using the \texttt{HuggingFace/datasets} library \citep{quentin-lhoest-datasets}.

\begin{table}[t]
\centering
\caption{Hyper-parameters used with Prune OFA}
\label{tab:hyper-pofa}
\begin{tabular}{@{}lc@{}}
\toprule
Hyper-parameter & Value \\ \midrule
Warmup ratio & 0.01 \\
Batch size & 256 \\
Weight decay & 0.01 \\
Max steps & 100k \\
Learning rate decay & Linear + LRR \\
Sequence length & 512 \\
$\lambda_{PT}$ & 0.5 \\
$\lambda_{kd}$ & 0.5 \\
Temperature & 2.0 \\
Pruning start & 0 \\
Pruning policy end & 50k \\
Pruning end & 80k \\
Pruning interval & 1k \\ \bottomrule
\end{tabular}%
\end{table}

\begin{table}[t]
\centering
\caption{Hyper-parameters used for transfer learning}
\label{tab:hyper-transfer}
\begin{tabular}{@{}lcc@{}}
\toprule
Hyper-parameter & SQuAD & GLUE \\ \midrule
Learning rate & \{1.5e-4, 1.8e-4\} & \{1e-4, 1.2e-4, 1.5e-5\} \\
Batch size & 12 & 32 \\
Weight decay & \multicolumn{2}{c}{\{0, 0.01\}} \\
Epochs & 8 & \{3, 6, 9\} \\
Learning rate decay & \multicolumn{2}{c}{Linear} \\
Warmup ratio & \multicolumn{2}{c}{\{0, 0.01, 0.1\}} \\
Sequence length & 384 & 128 \\
$\lambda_{PT}$ & \multicolumn{2}{c}{0.0} \\
$\lambda_{kd}$ & \multicolumn{2}{c}{1.0} \\
Temperature & \multicolumn{2}{c}{2.0} \\ \bottomrule
\end{tabular}%
\end{table}

\begin{table}[t]
\centering
\caption{Hyper-parameters used for quantization-aware training}
\label{tab:hyper-quant}
\begin{tabular}{@{}lcc@{}}
\toprule
Hyper-parameter & SQuAD & GLUE \\ \midrule
Learning rate & \{1e-6, 1e-5\} & \{5e-8, 1e-7, 1e-6, 1e-5\} \\
Batch size & 12 & 32 \\
Weight decay & \multicolumn{2}{c}{\{0, 0.01\}} \\
Epochs & 2 & 3 \\
Learning rate decay & \multicolumn{2}{c}{Linear} \\
Warmup ratio & \multicolumn{2}{c}{\{0, 0.01, 0.1\}} \\
Sequence length & 384 & 128 \\
$\lambda_{PT}$ & \multicolumn{2}{c}{0.0} \\
$\lambda_{kd}$ & \multicolumn{2}{c}{1.0} \\
Temperature & \multicolumn{2}{c}{2.0} \\ \bottomrule
\end{tabular}%
\end{table}

\subsection{Training details \& hyper-parameters}
\label{app:hyper-params}
\paragraph{Teacher preparation}
We execute the teacher preparation step on all models.
The pre-training objectives for both BERT models and DistilBERT are the same as in the original paper.
For BERT models, the objectives are masked language-modeling (MLM) and next sentence predicition (NSP), and for DistilBERT the objective is MLM only.
The hyper-parameters used are detailed in Table~\ref{tab:hyper-pofa}.
We use Adam optimizer \citep{kingma2015adam} with learning rates \{5e-5, 1e-4, 1e-4\} for \{BERT-Base, BERT-Large, DistilBERT\}.

\paragraph{Student pruning}
We run student pruning with the same objectives, hyper-parameters and optimizer we used at the teacher preparation step (Table~\ref{tab:hyper-pofa}) with learning rates \{1.5e-4, 1e-4, 1.5e-4\} for \{BERT-Base, BERT-Large, DistilBERT\}.

\paragraph{Transfer learning}
For transfer learning experiments of either Prune OFA or fine-tune pruning we use the hyper-parameters in Table~\ref{tab:hyper-transfer} coupled with Adam optimizer.
When combining knowledge distillation in the transfer learning phase, in our experiments we found that it is best to optimize only on knowledge distillation objective and ignore the ground truth labels.

\paragraph{Quantization}
For quantization-aware training experiments of Prune OFA  we use the hyper-parameters in Table~\ref{tab:hyper-quant} coupled with Adam optimizer.

\end{document}